%% file: force_ctrl_neurips.tex
\documentclass{article}

\usepackage[final]{neurips_2023}

\usepackage[utf8]{inputenc} 
\usepackage[T1]{fontenc}    
\usepackage{hyperref}       
\usepackage{url}            
\usepackage{booktabs}       
\usepackage{amsfonts}       
\usepackage{nicefrac}       
\usepackage{microtype}      
\usepackage{xcolor}         

\usepackage{graphicx}
\usepackage{siunitx}
\usepackage{subcaption}
\usepackage{amsmath}
\usepackage{xspace}
\usepackage{multirow}
\usepackage{multicol}

\usepackage{varwidth}
\newsavebox\tmpbox

\input{misc/math_commands}
\input{misc/more_math}

\title{Towards Transferring Tactile-based Continuous Force Control Policies from Simulation to Robot}

\author{%
  Luca Lach\\
  BarcelonaTech
  \And
  Robert Haschke \\
  Bielefeld University
  \And
  Davide Tateo \\
  TU Darmstadt \\
  \And
  Jan Peters \\
  TU Darmstadt \\
  \AND 
  Helge Ritter \\
  Bielefeld University  \\
  \And 
  Júlia Borràs \\
  BarcelonaTech \\
  \And
  Carme Torras \\
  BarcelonaTech
}

\begin{document}

\maketitle

\begin{abstract}
The advent of tactile sensors in robotics has sparked many ideas on how robots can leverage direct contact measurements of their environment interactions to improve manipulation tasks.
An important line of research in this regard is that of grasp force control, which aims to manipulate objects safely by limiting the amount of force exerted on the object.
While prior works have either hand-modeled their force controllers, employed model-based approaches, or have not shown sim-to-real transfer, we propose a model-free deep reinforcement learning approach trained in simulation and then transferred to the robot without further fine-tuning.
We therefore present a simulation environment that produces realistic normal forces, which we use to train continuous force control policies.
An evaluation in which we compare against a baseline and perform an ablation study shows that our approach outperforms the hand-modeled baseline and that our proposed inductive bias and domain randomization facilitate sim-to-real transfer.
Code, models, and supplementary videos are available on \url{https://sites.google.com/view/rl-force-ctrl}
\end{abstract}

\section{Introduction}

The application domain of tactile sensors is as diverse as the sensing principles within the field.
Complex sensors are commonly used in more challenging, high-level tasks such as surface following or edge prediction, which often involve deep learning methods (\cite{zhangRoboticCurvedSurface2020, dingSimtoRealTransferOptical2020, dingSimtoRealTransferRobotic2021, pengSimtoRealTransferRobotic2018, lachPlacingTouchingEmpirical2023}) or dexterous manipulation tasks (\cite{maoLearningFinePinchGrasp2023, melnikUsingTactileSensing2021}).
Low-level tasks like force control are commonly modeled by hand (\cite{romano2011human, tahara2010dynamic, li2012towards, lachBioInspiredGraspingController2022}), while those that use learning either rely on classical learning methods, do not investigate sim-to-real transfer, or both (\cite{perrusquiaPositionForceControl2019, luoReinforcementLearningVariable2019}).
In contrast, this paper presents a deep reinforcement learning (DRL) approach for the low-level task of grasp force control for parallel-jaw grippers with two degrees of freedom (DoFs) with two main control objectives: I) reaching and maintaining a given goal force, and II) minimizing object movements while closing and holding the object.


\begin{figure}[t]
\hfill
\begin{subfigure}[b]{0.285\linewidth}
   \includegraphics[width=1\linewidth]{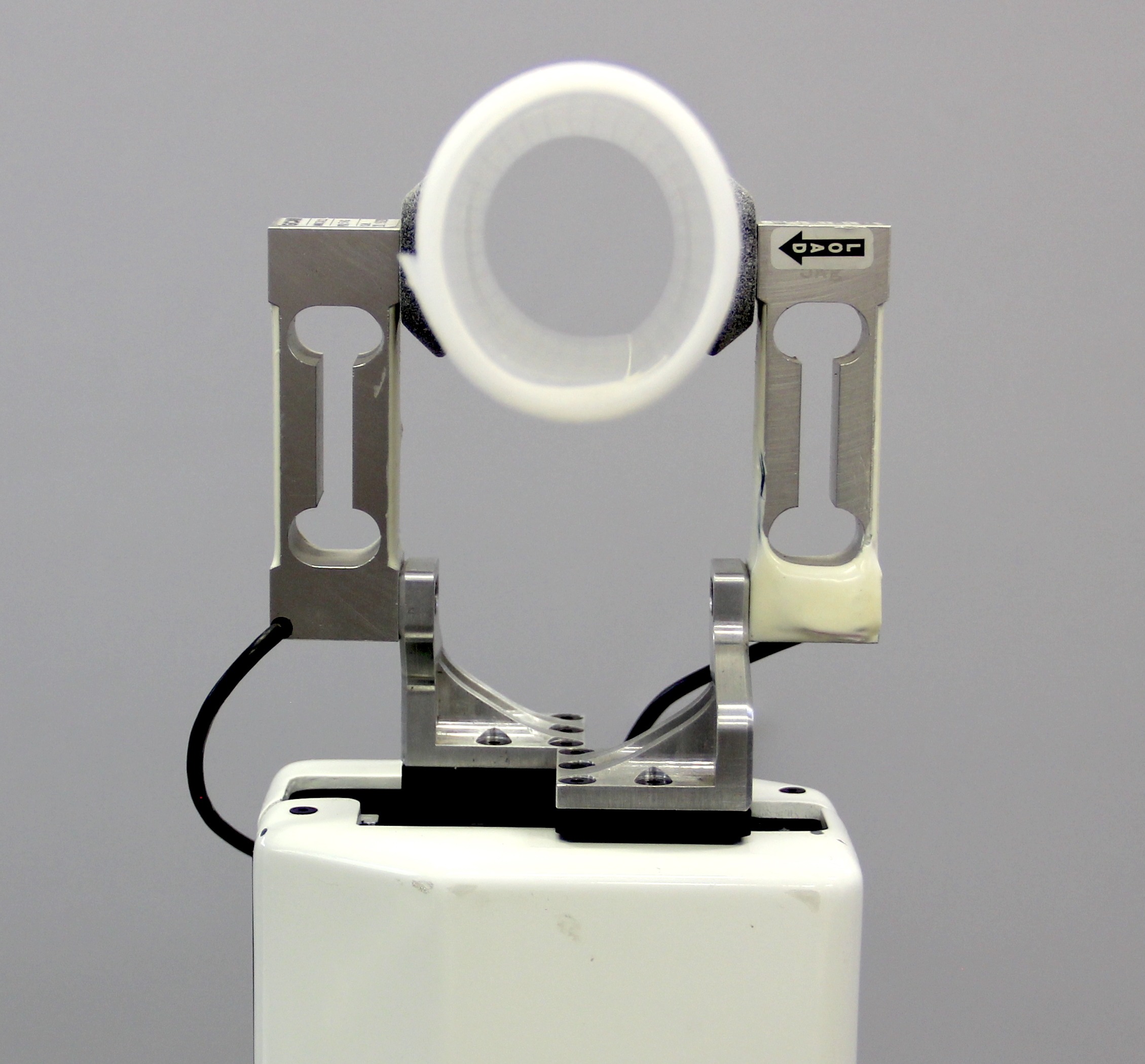}
   \caption{Real-life grasp}
   \label{fig:real_grasp} 
\end{subfigure}
\hfill
\begin{subfigure}[b]{0.325\linewidth}
   \includegraphics[width=1\linewidth]{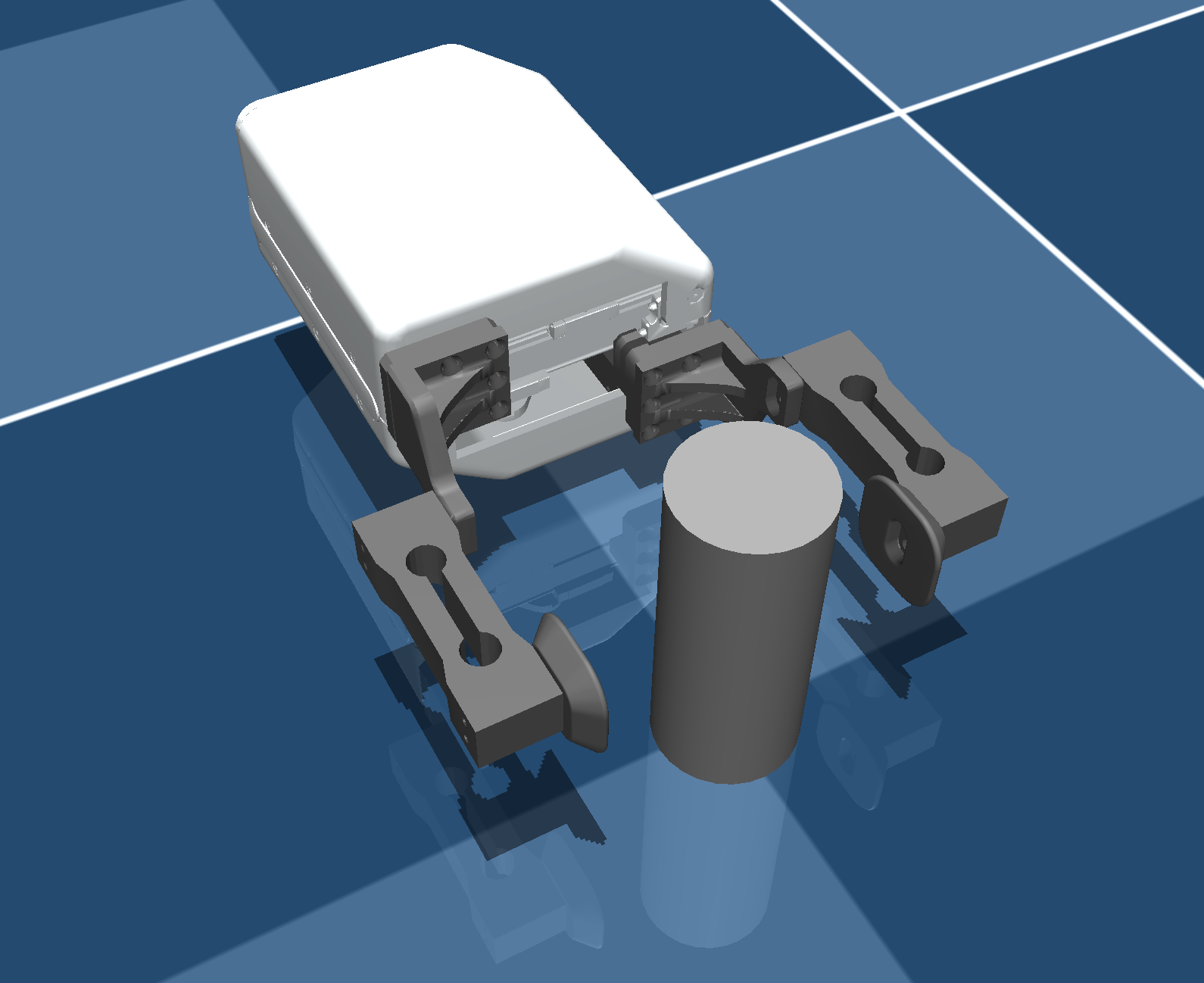}
   \caption{Simulation environment}
   \label{fig:sim_grasp} 
\end{subfigure}
\hfill
\begin{subfigure}[b]{0.34\linewidth}
\includegraphics[width=1\linewidth]{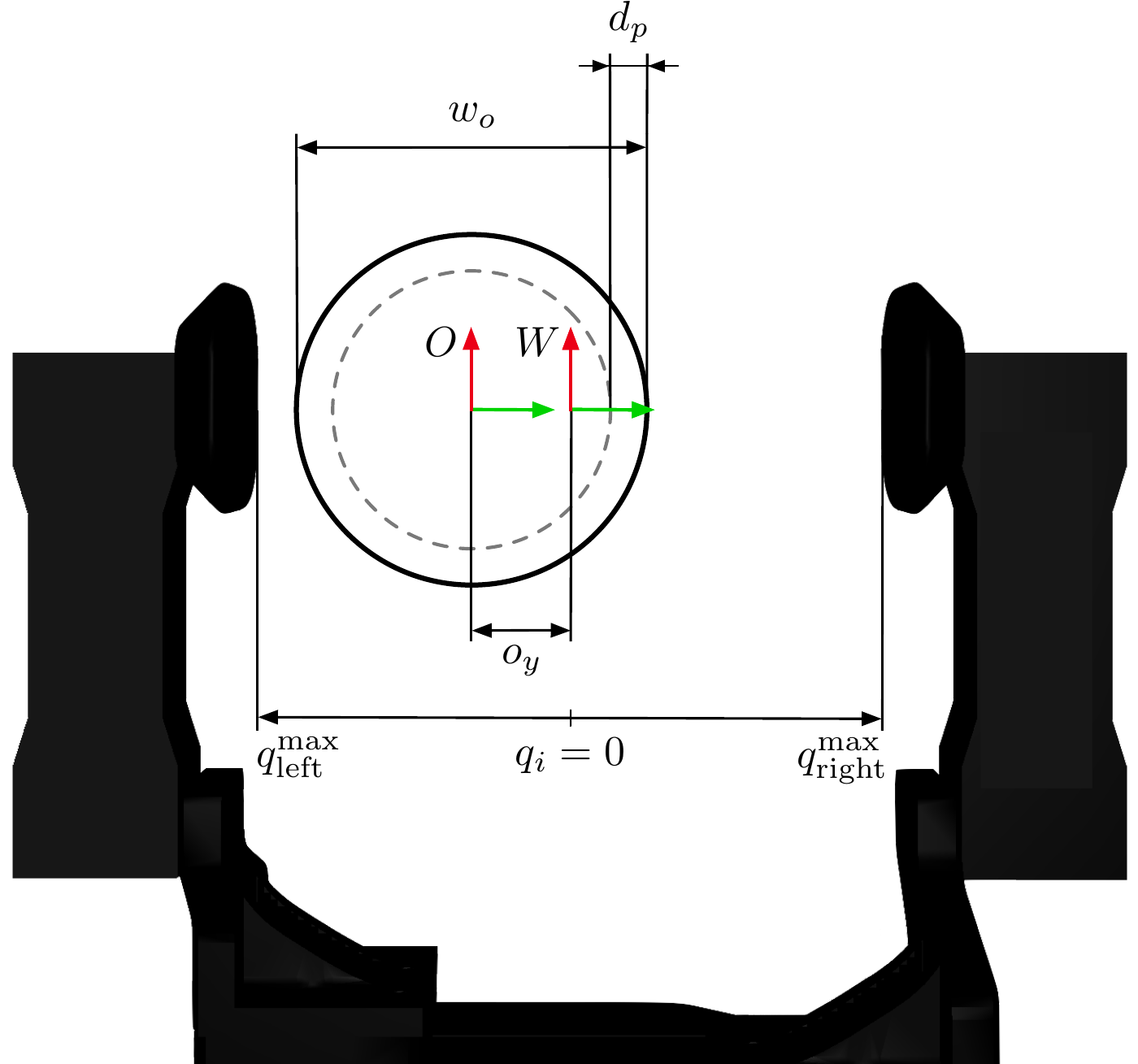}
\caption{Schematic overview}
\label{fig:gripper_scheme} 
\end{subfigure}
\hfill
\caption{Overview of the grasping scenario we consider in real life and simulation.} 
\label{fig:introFig}
\vspace{-0.7cm}
\end{figure}

In the following, we propose a simulation environment based on MuJoCo (\cite{todorovMuJoCoPhysicsEngine2012a}), where we tuned contact model parameters to match a few real-world samples.
Then, we detail our learning process based on deep reinforcement learning, where we apply domain randomization and introduce an inductive bias to learn policies for subsequent sim-to-real transfer.
Lastly, we compare our policy to a hand-modeled force controller from \cite{lachBioInspiredGraspingController2022}, and perform an ablation study on some model choices. 
Our main contributions are: i)
a training procedure based on reinforcement learning that generalizes zero-shot to the real robot, ii) a novel simulation environment for 2-DoF grippers with realistic fingertip forces, and iii) open-sourcing the code for the environment, all methods and their evaluation and CAD models of the sensorized gripper.
To the best of our knowledge, this is the first paper proposing a tactile-based continuous grasp force controller learned with DRL which was transferred to the real robot without further refinement.

\section{Related Work}

\noindent\textbf{Grasp Force Control} 
In their review of human grasping, \cite{johansson2009coding} highlight the importance of tactile sensations, and divide the human grasping sequence into distinct phases, where tactile events often mark phase transitions.
Many works from the robotics community presenting hand-modeled gripper controllers modeled their controllers accordingly (\cite{romano2011human, lachBioInspiredGraspingController2022, Hsiao2010, Patel2018}). 
In our work, we adapt this scheme through the inductive bias we apply to our policy actions. 
Learning grasp force control has also become popular in recent years. 
In \cite{merzicLeveragingContactForces2018a}, the authors learn a grasping policy that controls forces on rigid objects in simulation, while \cite{Wu2019} focuses on increasing grasp success using tactile feedback.
Others have learned to control grasping forces for more complex tasks like door opening (\cite{kangVersatileDoorOpening2023}), high-precision assembly tasks (\cite{luoReinforcementLearningVariable2019}), or surface tracking (\cite{zhangRoboticCurvedSurface2020}).
Although these works learn force control behaviors, none of them investigate the potential of learning them in simulation and transferring them to the real robot afterward.

\noindent\textbf{Sim-to-real transfer}
Sim-to-real transfer is widely used in robotics (\cite{zhaoSimtoRealTransferDeep2020, juTransferringPolicyDeep2022}) to avoid the time-consuming and labor-intensive task of real-world data collection.
In the domain of optical tactile sensors, \cite{churchTactileSimtoRealPolicy} and \cite{linTactileGymSimtoreal2022} presented Tactile Gym, a simulation environment containing the TacTip (\cite{ward-cherrierTacTipFamilySoft2018}), DIGIT (\cite{lambetaDIGITNovelDesign2020}) and DigiTac (\cite{leporaDigiTacDIGITTacTipHybrid2022}) sensors, and propose a domain adaptation approach by training adversarial networks on real-world data to generate tactile feedback in simulation. 
Approaches based on Finite Element Methods (FEM), are capable of generating accurate simulation data of complex tactile sensors by simulating their deformation (\cite{narangSimtoRealRoboticTactile2021, narangInterpretingPredictingTactile2021, sferrazzaDesignMotivationEvaluation2019, sferrazzaLearningSenseTouch2020, sferrazzaSimtorealHighresolutionOptical2021}.
Due to their high computational cost, FEM is typically not well-suited for data-driven approaches like DRL, unless some simplifying assumptions can be made (\cite{biZeroShotSimtoRealTransfer2021}).
In contrast to these studies, we focus on low-level force control tasks that solely require force measurements as inputs.
\cite{dingSimtoRealTransferRobotic2021} use MuJoCo to simulate a self-made tactile sensor array to open up a cabin door.
They again employ domain randomization for transferring the policy, but note that they binarized the sensor readings due to the low sensitivity of the built-in MuJoCo touch sensor.
Although \cite{akkaya2019solving} also mentions a potential lack of realism in simulated continuous force measurements, we find that continuous force control policies can indeed be learned in MuJoCo and then be successfully transferred to the real world without fine-tuning.

\section{Force Control Simulation}
\label{sec:env}

To train force control policies, we first modeled TIAGo's 2-DoF parallel jaw gripper in MuJoCo with one tactile sensor per finger and an object of variable softness.
The control frequency was set to \SI{25}{\Hz} to match that of a real TIAGo.
At each simulation step $t$, the environment executes $\qdes_i = q_i + u_i$, where $u_i = \dqdes_i$ is the control signal given by a policy or user.
It refers to the desired position delta, which is then added to the joint's current position and forwarded to the controllers.

\figref{fig:gripper_scheme} shows a schematic overview of the grasping scenario including all parameters required to define it.
The gripper is depicted in its fully open state ($q_i = q_i^\text{max} = 0.045$), with an object located between the fingers somewhere on the grasping axis.
$W$ and $O$ refer to the world and object frames, \pdep the maximum object deformation, \oy the object displacement, and \wo to the object width, where the two latter parameters are sampled upon episode initialization.

Next, we tune the simulation actuators and force sensors to be similar to the behavior of the real robot.
We also identify parameter ranges in which the simulation behaves realistic, yet different from our robot, so that our domain randomization will be highly diverse but not unrealistic.
To this end, we perform several grasps on the real robot, where we command different \dq (since TIAGo is position-controlled), which remain constant for each trial.
We then record the joint and force trajectories, repeat the experiments in simulation, and then tune the simulation parameters to match the real-world trajectories.
We use MuJoCo's solver impedance parameter \texttt{width} to model \dfdq, depending on the object stiffness, and introduce a scaling parameter \fscale that is multiplied with the simulation force to model lower forces for softer objects at the same position deltas.
In order to avoid unrealistic parameter combinations, e.g. soft objects with high \fscale, we introduce $\kappa \in [0,1]$, which we use to interpolate on the intervals of \texttt{width} $\in [0.003, 0.01]$ and $\fscale \in [0.5, 5]$.
Lastly, we found that the actuator bias parameter $b_2 \in [-13, -6]$  generates realistic motor behavior.

\section{Learning Methods}
\begin{figure}[t]
\hfill
\begin{subfigure}[b]{0.47\linewidth}
    \includegraphics[width=1\linewidth]{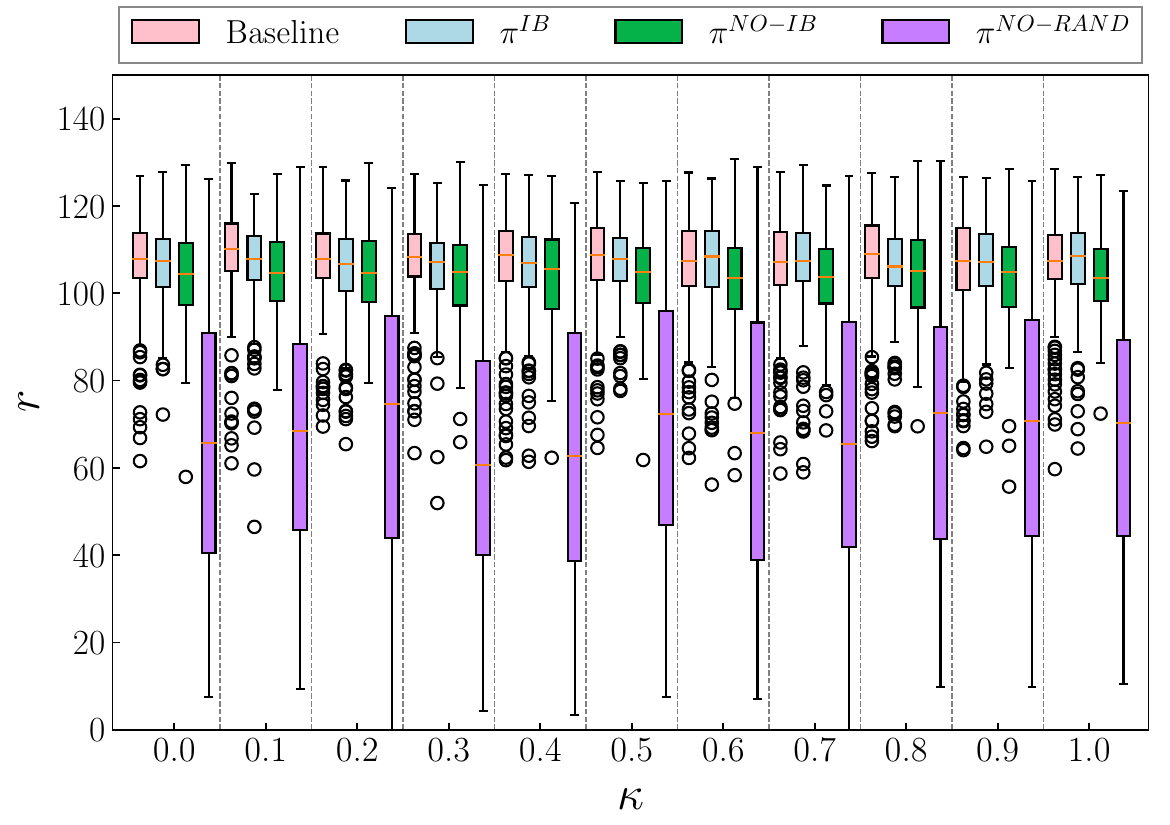}
    \caption{Comparison of rewards for different $\kappa$}
    \label{fig:box} 
\end{subfigure}
\hfill
\begin{subfigure}[b]{0.51\linewidth}
   \includegraphics[width=1\linewidth]{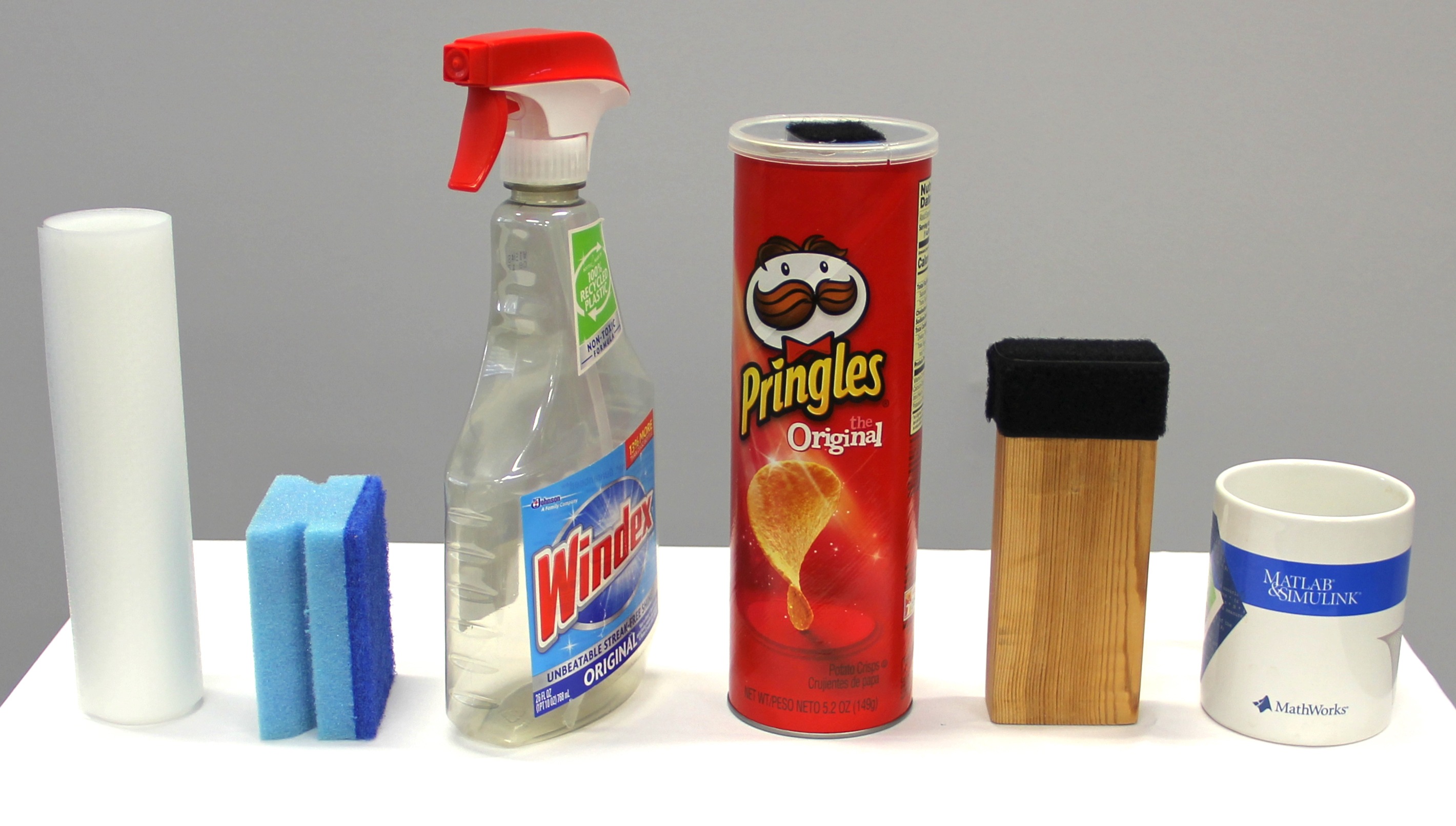}
   \caption{Evaluation objects ordered by stiffness}
   \label{fig:objects} 
\end{subfigure}
\hfill
\caption{Evaluation of our method in simulation (left) and on real-world objects (right).} 
\vspace{-0.5cm}
\end{figure}

As the task of force control is a sequential decision-making problem, we can model it as a Markov Decision Process (MDP), allowing us to solve the problem using RL algorithms.
The agent receives the following observation at each time step $o(t) =
\left ( q_i, f_i, \df_i, a_i(t-1), h_i \right )^T$, where subscript $i$ indicates that the observation is given for all joints $i$, $\df_i = \fgoal - f_i$  refers to the difference of the current force to the goal force, $a_i$ the last action taken by the agent, and $h_i$ being the \texttt{had\_contact} flag, which switches from 0 to 1 upon contact acquisition and stays 1 even if contact is lost afterward.
We add gaussian noise to the joint position and fingertip force and stack the observation $k=3$ times.
The action space is simply a two-vector with one desired position per finger $(a_\text{left}, a_\text{right})^T$, where $a_i = \dqdes_i$ refers to an individual action for joint $i$.
Each $a_i$ is first clipped to lie within $[-1, 1]$ and then multiplied with \dqmax, effectively denormalizing $a_i$.
Additionally, we define a contact-state dependent inductive bias $\phi_i$ that acts as a scaling factor for the individual actions at each time step, yielding $a_i' = \phi_i \, a_i$.
Thereby, our policy imitates the human grasping phases from~\cite{johansson2009coding} and is safer to execute on the real robot as erratic joint movements are less likely.

Our reward function contains three terms, one for each controller objective and another one to foster a smooth control behavior.
We give a continuous reward that is highest when $\df=0$, a sparse reward that penalizes object movements with -1, and we additionally penalize the difference between the last and current action.
Finally, all individual rewards are weighted and a single, scalar reward is computed per time step.
The heavy movement penalty will be given frequently during the early, exploratory phase of training.
Therefore, we employ a learning curriculum that increases the penalty weight and amount of domain randomization during training, thereby increasing environment complexity.

\section{Experimental Evaluation}

\begin{table*}[t]
\begin{center}
\caption{\small Real-world experimental results for six household objects. Our method \polib performs strongest, closely followed by the baseline. The ablated policies showed weaker sim-to-real transfer.} 
\label{table:real_eval}
\scriptsize
\scalebox{1.0}{
\begin{tabular}{lcccccccc}

Model & Metric & Rubber Mat & Sponge & Spray & Pringles & Wood & Mug & Average\\
\toprule

\multirow{2}{*}{Baseline} & 
    \multirow{2}{*}{\shortstack{ Reward \\ Obj.Mov. }} &
    \multirow{2}{*}{\shortstack{ $92 \pm 15$  \\ $1.6 \pm 0.8$   }} & 
    \multirow{2}{*}{\shortstack{ $96 \pm 14$  \\ $\textbf{3.1} \pm \textbf{2.1}$   }} & 
    \multirow{2}{*}{\shortstack{ $\textbf{114} \pm \textbf{5}$  \\ $1.9 \pm 0.6$   }} & 
    \multirow{2}{*}{\shortstack{ $\textbf{118} \pm \textbf{7}$  \\ $\textbf{0.9} \pm \textbf{0.4}$   }} & 
    \multirow{2}{*}{\shortstack{ $\textbf{108} \pm \textbf{7}$  \\ $1.0 \pm 0.9$   }} & 
    \multirow{2}{*}{\shortstack{ $115 \pm 10$  \\ -   }} &
    \multirow{2}{*}{\shortstack{ $107 \pm 10$  \\ $\textbf{1.7} \pm \textbf{0.9}$   }} \\
&&&&&&&&\\
\midrule

\multirow{2}{*}{\polib} & 
    \multirow{2}{*}{\shortstack{ Reward \\ Obj.Mov. }} &
    \multirow{2}{*}{\shortstack{ $\textbf{96} \pm \textbf{15}$  \\ $\textbf{1.5} \pm \textbf{0.8}$   }} & 
    \multirow{2}{*}{\shortstack{ $\textbf{103} \pm \textbf{7}$  \\ $3.2 \pm 1.4$   }} & 
    \multirow{2}{*}{\shortstack{ $109 \pm 17$  \\ $\textbf{1.8} \pm \textbf{0.8}$   }} & 
    \multirow{2}{*}{\shortstack{ $115 \pm 16$  \\ $1.2 \pm 0.6$   }} & 
    \multirow{2}{*}{\shortstack{ $105 \pm 15$  \\ $\textbf{0.7} \pm \textbf{0.7}$   }} & 
    \multirow{2}{*}{\shortstack{ $\textbf{120} \pm \textbf{10}$  \\ -   }} &
    \multirow{2}{*}{\shortstack{ $\textbf{109} \pm \textbf{13}$  \\ $\textbf{1.7} \pm \textbf{0.9}$  }} \\
&&&&&&&&\\
\midrule

\multirow{2}{*}{\polnoib} & 
    \multirow{2}{*}{\shortstack{ Reward \\ Obj.Mov. }} &
    \multirow{2}{*}{\shortstack{ $77 \pm 10$  \\ $2.8 \pm 1.4$   }} & 
    \multirow{2}{*}{\shortstack{ $74 \pm 26$  \\ $4.0 \pm 0.9$   }} & 
    \multirow{2}{*}{\shortstack{ $88 \pm 33$  \\ $2.4 \pm 0.7$   }} & 
    \multirow{2}{*}{\shortstack{ $98 \pm 34$  \\ $1.9 \pm 0.6$   }} & 
    \multirow{2}{*}{\shortstack{ $101 \pm 16$  \\ $1.7 \pm 0.4$   }} & 
    \multirow{2}{*}{\shortstack{ $106 \pm 17$  \\ -   }} &
    \multirow{2}{*}{\shortstack{ $89 \pm 23$  \\ $2.6 \pm 0.8$    }} \\
&&&&&&&&\\
\midrule

\multirow{2}{*}{\polnorand} & 
    \multirow{2}{*}{\shortstack{ Reward \\ Obj.Mov. }} &
    \multirow{2}{*}{\shortstack{ $82 \pm 12$  \\ $2.6 \pm 1.6$   }} & 
    \multirow{2}{*}{\shortstack{ $97 \pm 15$  \\ $3.6 \pm 1.3$   }} & 
    \multirow{2}{*}{\shortstack{ $57 \pm 37$  \\ $2.6 \pm 0.9$   }} & 
    \multirow{2}{*}{\shortstack{ $53 \pm 39$  \\ $1.7 \pm 0.7$   }} & 
    \multirow{2}{*}{\shortstack{ $49 \pm 33$  \\ $1.8 \pm 0.9$   }} & 
    \multirow{2}{*}{\shortstack{ $40 \pm 41$  \\ -   }} &
    \multirow{2}{*}{\shortstack{ $66 \pm 30$  \\ $2.5 \pm 1.1$    }} \\
&&&&&&&&\\
\bottomrule 
\end{tabular}}
\end{center}
\vspace{-0.6cm}
\end{table*}

For the experiments, we use Proximal Policy Optimization (PPO) from \cite{schulmanProximalPolicyOptimization2017} to train the grasping policies.
We first evaluate our proposed method in simulation, and then apply the policies to the real robot and compare them in terms of force reward and object movements.
We compare our policy with inductive bias \polib to a Python implementation of the grasp force controller presented in~\cite{lachBioInspiredGraspingController2022}, a policy \polnoib trained without the inductive bias, and a policy \polnorand trained with neither inductive bias nor domain randomization.
We evaluated both models on 11 stiffness values $\kappa$ (spaced out evenly in $[0,1]$, including interval borders) for 200 simulation trials each, summing up to 2200 trials per model and 8800 trials in total.
Note, that during these trials all other environment parameters except $\kappa$ were randomly sampled for each trial.
\Figref{fig:box} shows a box plot of cumulative episode rewards for each model at each value for $\kappa$.
It shows that all models using domain randomization can successfully control grasping forces while minimizing object movements.
\polnorand however performed significantly worse, even on $\kappa = 0.5$ which it was trained on since actuator parameters were varied during evaluation while they were fixed during training.

To assess whether our policies are general enough to be transferred to the real robot, we evaluate them on TIAGo using six test objects of varying stiffness (see \figref{fig:objects}), performing  20 grasping trials per object and method, yielding $6 \times 4 \times 20 = 480$ trials in total.
In each trial, the object is offset to one finger, and the total object displacement is measured afterward as we did not have access to object velocity measurements.
The force reward was calculated identically to the simulation, but since the object velocity is not included, the rewards between real and simulation are not directly comparable.
\Tabref{table:real_eval} shows the results of the real-world evaluation, where \polib shows the strongest overall performance, slightly outperforming the baseline.
\polnoib and \polnorand both exhibit larger object movements, showing the importance of the inductive bias. 
\polnorand performed poorly in terms of reward, which shows that domain randomization is a crucial component for sim-to-real transfer.
The evaluation clearly shows that domain randomization is crucial for successful zero-shot policy transfer, and that domain knowledge in the form of inductive biases further facilitates the transfer.
Our proposed simulation environment has shown to generate realistic forces, such that the transfer was possible for continuous control policies.

\section{Conclusion}

In this work, we presented a DRL method to train grasp force controllers for 2-DoF grippers in simulation and subsequently transfer them to the real robot without fine-tuning.
We proposed a novel simulation environment that generates realistic grasp forces, which we used to train our policies.
To strengthen the transfer performance, we proposed to use an inductive bias and domain randomization.
An extensive real-world evaluation has shown that our method can successfully grasp objects of highly varying stiffnesses while minimizing object movements during the grasp. 
Conclusively, our results show that continuous force control policies can be learned end-to-end is simulation and slightly outperform hand-modeled controllers on real robots.
An interesting direction for future work is the integration of grasp force control in more complex tasks for which DRL methods are used.

\medskip


\section*{Acknowledgments}

This work was supported by the European Union's Horizon 2020 Marie Curie Actions under grant no. 813713 NeuTouch, the Horizon Europe research and innovation program under grant no. 101070600 SoftEnable, the German Research Center for AI (DFKI), and Hessian.AI. Research presented in this paper has been supported by the German Federal Ministry of Education and Research (BMBF) within the collaborative KIARA project (grant no. 13N16274).

\bibliography{force_ctr_neurips}

\clearpage

\section*{Appendix}

\subsection*{Force Control Simulation}

\begin{figure*}[b!]
\hfill
\begin{subfigure}[b]{0.325\linewidth}
   \includegraphics[width=1\linewidth]{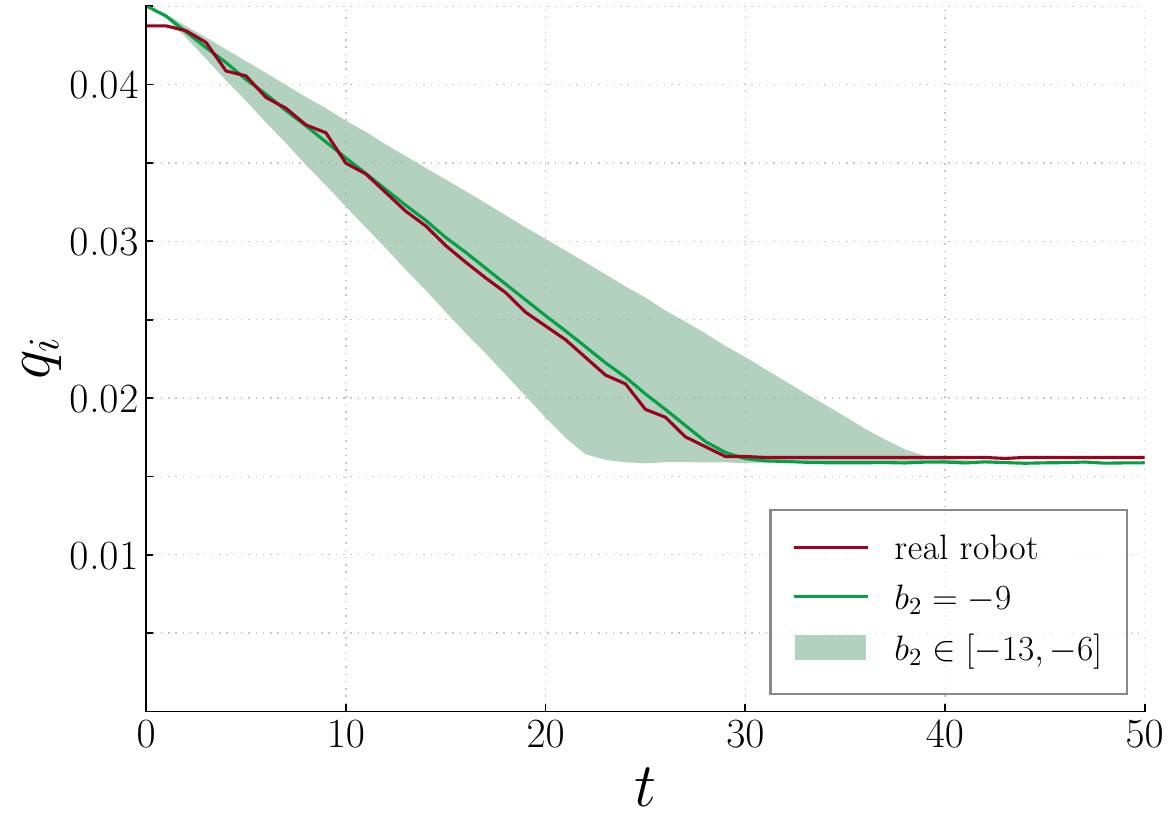}
   \caption{Variation of $b_2$, changing finger velocity}
   \label{fig:act_var} 
\end{subfigure}
\hfill
\begin{subfigure}[b]{0.325\linewidth}
   \includegraphics[width=1\linewidth]{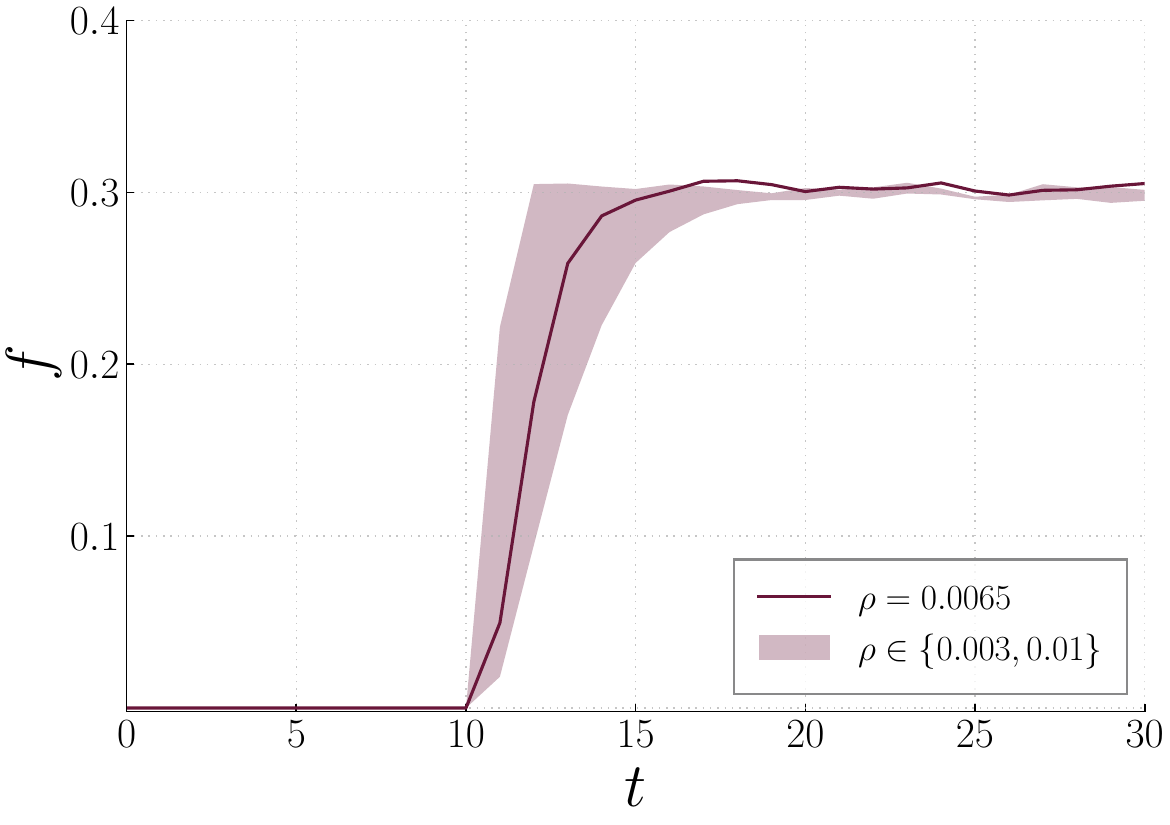}
   \caption{Variation of $\rho$, changing \dfdq}
   \label{fig:solimp_var} 
\end{subfigure}
\hfill
\begin{subfigure}[b]{0.325\linewidth}
   \includegraphics[width=1\linewidth]{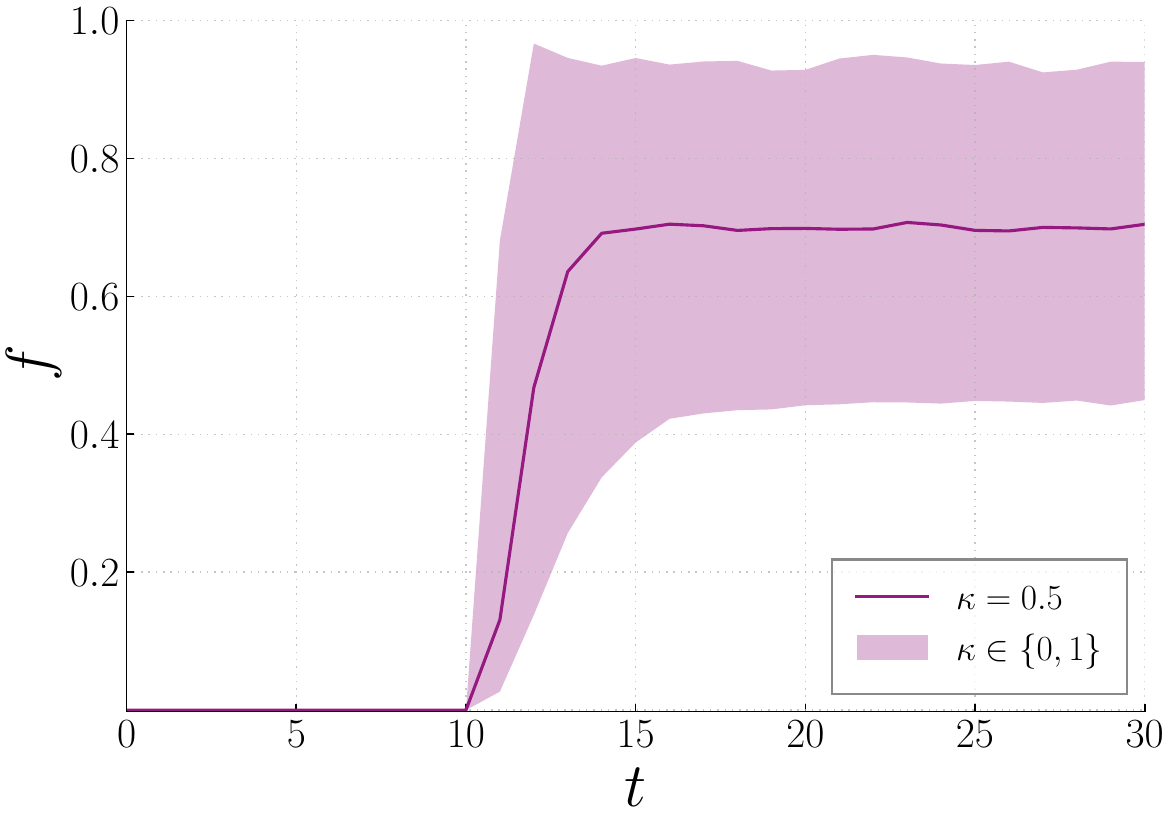}
   \caption{Variation of $\kappa$, changing \dfdq and $f(T)$}
   \label{fig:fscale_var} 
\end{subfigure}
\hfill
\caption{\small Results from the real-world grasping trials conducted to determine realistic simulation parameters. Each plot shows the effect of varying one of the parameters discussed in \Secref{sec:env}, where $b_2$ changes actuator behavior and $\kappa$ affects the forces by changing $\rho$ and $\fscale$.}
\vspace{-0.4cm}
\end{figure*}

The following three constraints are imposed on the sampling of the environment parameters from \figref{fig:gripper_scheme}:
\begin{align}
|\oy| + r_o &< \qmax \label{eq:const_qmax} \\
r_o -\pdep  &> |\oy| \label{eq:const_oy} \\
\pdep &< r_o \label{eq:const_wodp}
\end{align}
where $r_o = \frac{1}{2}\wo$ is the object radius.
The joints are controlled by
\begin{align}
    \qdes_i = q_i + u_i
\end{align}
at each simulation step $t$, where $u_i = \dqdes_i$ is the control signal the policy or a user passes to the environment.
Using position deltas instead of positions directly has several advantages for the learning process.
First, the action space is symmetric and centered around zero with known bounds, making it easy to normalize while satisfying assumptions some RL algorithms make about the action space.
Second, by constraining the control signals with $|u_i| \leq \dqmax$, the maximum joint velocity is bounded and prevents the policy from executing erratic and potentially dangerous movements.
Choosing \dqmax is straightforward by executing different gripper closing trials with increasing values for \dqmax and setting it to the value resulting in the highest velocity within safety limits.
In all experiments, we set $\dqmax = 0.003$.

After defining the control, we tuned the actuators in MuJoCo to match the behavior of TIAGo. 
First, \textit{gainprm} and \textit{ctrlrange} were set to $[0.0, 0.045]$ and $[100, 0, 0]$ to mimic the real joint actuation range $[0, 0.045]$, which reflects the desired finger position in centimeters (see \figref{fig:gripper_scheme}).
Then, the third parameter of \textit{biasprm}, $b_2$, was tuned to match the finger's closing velocity and acceleration.
In order to find a range of realistic values for $b_2$, we executed power grasps with the real robot and in simulation, and then manually tuned $b_2$ until the joint trajectories matched.
We found $b_2 = -9$ to mirror the real behavior best, and $b_2 \in [-13, -6]$ to result in realistic actuator behavior for our domain randomization, which is shown in \figref{fig:act_var}.

After identifying realistic actuator parameters, the contact forces need to be modeled to imitate real-world objects as well. 
In MuJoCo, the softness of contact constraints can be changed with the \textit{solimp} parameters, which allow more object penetration, resulting in different \dqdf.
We set  \textit{solimp}'s first parameters, $d_\text{min} = 0$ to allow constraints to be maximally soft and use its \textit{width} parameter, which we refer to as $\rho$, to vary the constraint softness for each trial.

In order to find a range of realistic values for $\rho$, we executed real-life grasping trials on a soft and a rigid object (Sponge and Wood from \figref{fig:objects}).
Then, we conducted the same experiment in simulation, compared \dfdq for both experiments, and tuned $\rho$ until the changes in force w.r.t. the joint position matched the real-world trials for both objects.
\Figref{fig:solimp_var} shows force trajectories for the determined interval $\rho \in [0.003, 0.01]$.

\begin{figure}[t]
\centering
\includegraphics[width=0.7\linewidth]{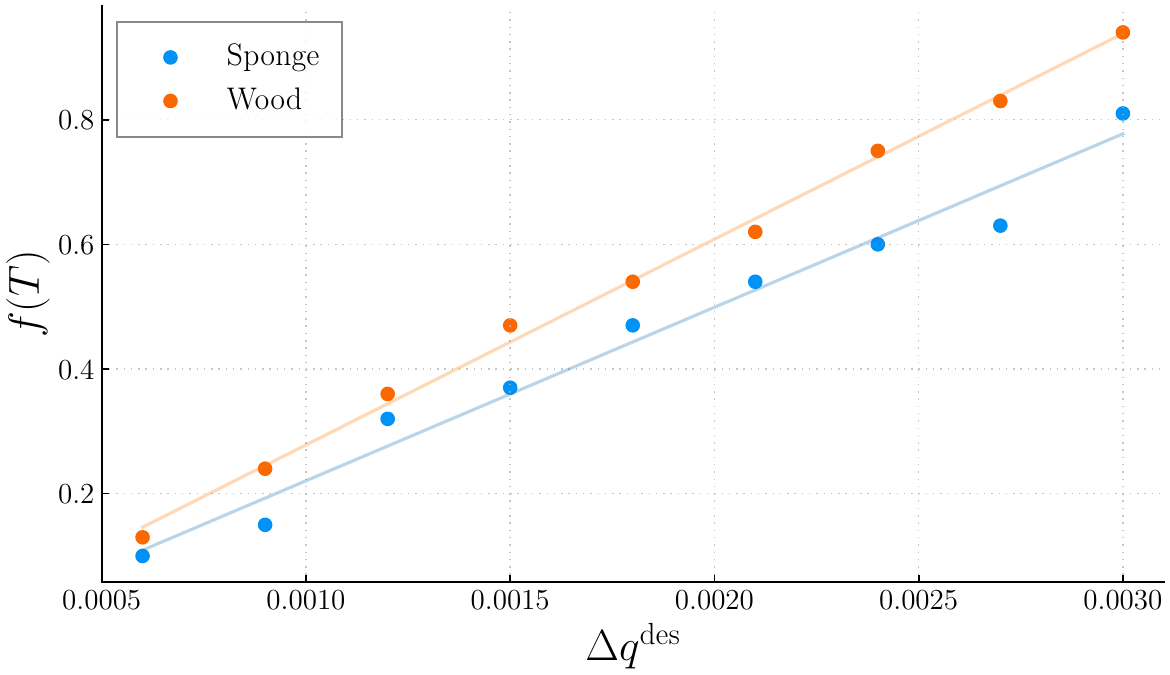}
\caption{\small Comparison of final forces $f(T)$ of Sponge and Wood for grasps with different \dqdes. The regression slopes are 278 and 330. }
\label{fig:stiffness_plot} 
\vspace{-0.4cm}
\end{figure}

Note, that changing the constraint stiffness $\rho$ results in different \dfdq, but not in different final forces.
In reality, however, the force exerted on a stiff object is higher than one exerted on a soft object for the same \dqdes.
To also model this behavior, we introduced a force scaling factor \fscale, such that stiffer objects generate higher forces more quickly, and softer objects slower.
We conducted another set of grasping trials on the same objects, where a small, constant \dqdes was commanded to both fingers and after the object was being held for a short amount of time, the final force $f(T)$ was noted.
The gripper was then opened again, \dqdes increased by $3 \times 10^{-4}$ and the experiment repeated until \dqdes = \dqmax.
Then, we repeated the same experiment in simulation and regressed $f(T)$ to \dqdes for both experiments.
The experiment data from the robot trials are shown in \figref{fig:stiffness_plot}.
\fscale is then the factor needed to match the regression slope of the simulation experiments with that of the real robot.
This way, we determined $\fscale \in [0.5, 5]$.

Lastly, to determine both values for $\rho$ and \fscale, we define a unified stiffness factor $\kappa \in [0, 1]$ which we use to interpolate within their respective intervals.
This way, we prevent unrealistic object configurations, e.g.\ a soft object with a high \fscale.

\subsection*{Learning Methods}

The \texttt{had\_contact} flag is defined as:
\begin{align}
  c_i(t) &= \begin{cases} 1 & \text{if}\,\, f_i(t) > f_\theta\\
    0 & \text{otherwise}
  \end{cases}\\
  h_i(t) &= c_i(t) \lor h_i(t-1)
\end{align}
where $c_i(t)$ indicates whether a finger is considered to be in contact with the object at time $t$ by comparing the current force to a noise threshold $f_\theta$, and $h_i(0) = 0$.
$h_i$ is 1 once $c_i$ was 1 before within the episode, even if the finger lost contact again ($c_i = 0$).
We provide this flag to avoid having a recurrent policy (since they can be difficult to train) or a long history of observations.
We add gaussian noise to the joint position and fingertip force with $\sigma_q = 0.000027$ and $\sigma_f = 0.013$ and stack the observation $k=3$ times for the policy to have access to a short history of position and force deltas so that it can estimate the object stiffness.

Additionally, we define a contact-state dependent inductive bias:
\begin{align}
    \phi_i = \begin{cases}
        \max(0.9, 1-\frac{|\df_i|}{\fgoal}) & \text{if} \,\, h_i = h_j = 1 \\
        0.1 & \text{if} \,\, h_i = 1 \land h_i \neq h_j \\
        1 & \text{else}
    \end{cases} 
\label{eq:indb}
\end{align}
which mimics the human grasping phases \cite{johansson2009coding} and is commonly used in other controllers (\cite{romano2011human, lachBioInspiredGraspingController2022}).

Our reward function mainly reflects the two controller objectives and adds a third term for smoother control.
We propose the following individual reward terms:
\begin{align}
  \label{eq:rew1}
  \rforce &= 1 - \tanh\left (\ \sum_i |\Delta f_i|\right )\\ 
 \label{eq:rew2}
  \robj &=  \begin{cases} -1 & \text{if}\,\, \oydot > \oydotmax \\ 
                         0 & \text{otherwise}
             \end{cases} \\  
  \label{eq:rew3}
  \ract &= - \sum_i | a_i(t-1) - a_i(t) | 
\end{align}
The term \rforce reflects the first control objective in the reward function, namely to reach and maintain the target force.
$\tanh$ normalizes the force delta in $[0,1]$ and subtracting it from 1 yields the highest reward if $\sum\df_i = 0$.
\robj expresses the second controller objective as a reward function by penalizing object movements with a sparse reward of -1 upon constraint violation.
In equation \ref{eq:rew2},  \oydot refers to the current object velocity and \oydotmax to the object velocity threshold.
\oydot has shown to always be noisy during finger contact, hence we use the threshold $\oydotmax = 0.00005$.
The third reward term, \ract, penalizes high changes in policy actions at consecutive time steps to encourage a smooth control behavior.
Other, more involved procedures have been proposed before (\cite{mysoreRegularizingActionPolicies2021}), but for our purposes this simple constraint has shown to be sufficient.
Finally, the total reward per time step is defined as
\begin{align}
    r = \alpha_1 \rforce + \alpha_2 \robj + \alpha_3 \ract
    \label{eq:totalr}
\end{align}

Our reward function as it is given in \eqref{eq:totalr} poses a difficult problem for learning control policies:
during exploration, it is highly likely that a random agent will push the object with one finger before it learns to control the grasping force precisely.
As a result, the policy will converge to local minima where it avoids contact with the object since it will receive a large negative reward from \robj before being rewarded by \rforce.
We, therefore, employ a learning curriculum, a common approach to gradually increase the task complexity over the course of training by annealing certain environment parameters (\cite{narvekarLearningCurriculumPolicies2018, narvekarCurriculumLearningReinforcement2020}).

The parameters controlling the likelihood of these high negative rewards are $\alpha_2$, the scaling factor for \robj, the range of possible object displacements \OY, the range of object radii \WO, and the object velocity threshold \oydotmax.
For the scalar parameter $\alpha_2$ and \oydotmax, a starting value and a final value are defined.
During training, the parameters are linearly interpolated at each time step from their initial value at $t=0$ until they reach their final value at $t=s_\text{end}$, where $s_\text{end}$ is a hyperparameter.
For the intervals \OY and \WO, the initial and final interval borders are given. 
During the annealing phase, both borders are also linearly interpolated from their respective initial to their final values.
All annealed parameters and their initial and final values are shown in \Tabref{tab:params}.

\begin{table}
\begin{center}
    \begin{tabular}{lcc}
     & Initial & Final  \\
    \toprule
    $\alpha_2$  & 0 & 1.0  \\
    \oydotmax   & $2 \times 10^{-4}$ & $5 \times 10^{-5}$  \\
    \WO         & $[0.020, 0.025]$ & $[0.015, 0.035]$  \\
    \OY         & $[0.0, 0.0]$ & $[-0.040, 0.040]$  \\
    \bottomrule
    \end{tabular}
    \vspace{0.2cm}
    \captionof{table}
      {%
        Annealed parameters with their initial and final values.
        \label{tab:params}%
      }
\end{center}
\vspace{-0.4cm}
\end{table}

\end{document}

%% file: misc/math_commands.tex
\def\figref#1{figure~\ref{#1}}
\def\Figref#1{Fig.~\ref{#1}}

\def\Secref#1{Section~\ref{#1}}

\def\1{\bm{1}}

\DeclareMathAlphabet{\mathsfit}{\encodingdefault}{\sfdefault}{m}{sl}
\SetMathAlphabet{\mathsfit}{bold}{\encodingdefault}{\sfdefault}{bx}{n}

\makeatletter
\newcommand{\StatexIndent}[1][3]{%
  \setlength\@tempdima{\algorithmicindent}%
  \Statex\hskip\dimexpr#1\@tempdima\relax}
\makeatother

%% file: misc/more_math.tex
\def\figref#1{Fig.~\ref{#1}}
\def\Figref#1{Fig.~\ref{#1}}

\def\Tabref#1{Table~\ref{#1}}

\def\eqref#1{equation~\ref{#1}}

\def\Secref#1{Sec.~\ref{#1}}

\newcommand{\fgoal}{\ensuremath{f^\text{goal}}\xspace}

\newcommand{\fscale}{\ensuremath{f_\alpha}\xspace}
\newcommand{\dfdq}{\ensuremath{\frac{\partial f}{\partial q}}\xspace}
\newcommand{\qmax}{\ensuremath{q^\text{max}}\xspace}
\newcommand{\qdes}{\ensuremath{q^\text{des}}\xspace}
\newcommand{\dq}{\ensuremath{\Delta q}\xspace}
\newcommand{\df}{\ensuremath{\Delta f}\xspace}
\newcommand{\dqdf}{\ensuremath{\frac{\partial f}{\partial q}}\xspace}
\newcommand{\dqdes}{\ensuremath{\Delta \qdes}\xspace}
\newcommand{\dqmax}{\ensuremath{\Delta \qmax}\xspace}

\newcommand{\rforce}{\ensuremath{r^\text{force}}\xspace}
\newcommand{\robj}{\ensuremath{r^\text{obj}}\xspace}

\newcommand{\ract}{\ensuremath{r^\text{act}}\xspace}
\newcommand{\wo}{\ensuremath{w_o}\xspace}
\newcommand{\pdep}{\ensuremath{d_p}\xspace}
\newcommand{\oy}{\ensuremath{o_y}\xspace}
\newcommand{\oydot}{\ensuremath{\dot{o}_y}\xspace}
\newcommand{\oydotmax}{\ensuremath{\dot{o}_y^\text{max}}\xspace}

\newcommand{\WO}{\ensuremath{W_o}\xspace}
\newcommand{\OY}{\ensuremath{O_y}\xspace}

\newcommand{\polib}{\ensuremath{\pi^\text{IB}}\xspace}
\newcommand{\polnoib}{\ensuremath{\pi^\text{NO-IB}}\xspace}
\newcommand{\polnorand}{\ensuremath{\pi^\text{NO-RAND}}\xspace}

%% file: force_ctrl_neurips.bbl
\begin{thebibliography}{38}
\providecommand{\natexlab}[1]{#1}
\providecommand{\url}[1]{\texttt{#1}}
\expandafter\ifx\csname urlstyle\endcsname\relax
  \providecommand{\doi}[1]{doi: #1}\else
  \providecommand{\doi}{doi: \begingroup \urlstyle{rm}\Url}\fi

\bibitem[Zhang et~al.(2020)Zhang, Xiao, Zou, Xiao, and
  Chen]{zhangRoboticCurvedSurface2020}
Tie Zhang, Meng Xiao, Yan-biao Zou, Jia-dong Xiao, and Shou-yan Chen.
\newblock Robotic {{Curved Surface Tracking}} with a {{Neural Network}} for
  {{Angle Identification}} and {{Constant Force Control}} based on
  {{Reinforcement Learning}}.
\newblock \emph{International Journal of Precision Engineering and
  Manufacturing}, 21\penalty0 (5):\penalty0 869--882, 2020.
\newblock ISSN 2234-7593, 2005-4602.

\bibitem[Ding et~al.(2020)Ding, Lepora, and
  Johns]{dingSimtoRealTransferOptical2020}
Zihan Ding, Nathan~F. Lepora, and Edward Johns.
\newblock Sim-to-{{Real Transfer}} for {{Optical Tactile Sensing}}, 2020.

\bibitem[Ding et~al.(2021)Ding, Tsai, Lee, and
  Huang]{dingSimtoRealTransferRobotic2021}
Zihan Ding, Ya-Yen Tsai, Wang~Wei Lee, and Bidan Huang.
\newblock Sim-to-{{Real Transfer}} for {{Robotic Manipulation}} with {{Tactile
  Sensory}}, 2021.

\bibitem[Peng et~al.(2018)Peng, Andrychowicz, Zaremba, and
  Abbeel]{pengSimtoRealTransferRobotic2018}
Xue~Bin Peng, Marcin Andrychowicz, Wojciech Zaremba, and Pieter Abbeel.
\newblock Sim-to-{{Real Transfer}} of {{Robotic Control}} with {{Dynamics
  Randomization}}.
\newblock In \emph{2018 {{IEEE International Conference}} on {{Robotics}} and
  {{Automation}} ({{ICRA}})}, pages 3803--3810, 2018.

\bibitem[Lach et~al.(2023)Lach, Funk, Haschke, Lemaignan, Ritter, Peters, and
  Chalvatzaki]{lachPlacingTouchingEmpirical2023}
Luca Lach, Niklas Funk, Robert Haschke, Severin Lemaignan, Helge~Joachim
  Ritter, Jan Peters, and Georgia Chalvatzaki.
\newblock Placing by {{Touching}}: {{An}} empirical study on the importance of
  tactile sensing for precise object placing.
\newblock In \emph{{{IROS23}}}, 2023.

\bibitem[Mao et~al.(2023)Mao, Xu, Wen, Kasaei, Yu, Psomopoulou, Lepora, and
  Li]{maoLearningFinePinchGrasp2023}
Xiaofeng Mao, Yucheng Xu, Ruoshi Wen, Mohammadreza Kasaei, Wanming Yu, Efi
  Psomopoulou, Nathan~F. Lepora, and Zhibin Li.
\newblock Learning {{Fine Pinch-Grasp Skills}} using {{Tactile Sensing}} from
  {{Real Demonstration Data}}, 2023.

\bibitem[Melnik et~al.(2021)Melnik, Lach, Plappert, Korthals, Haschke, and
  Ritter]{melnikUsingTactileSensing2021}
Andrew Melnik, Luca Lach, Matthias Plappert, Timo Korthals, Robert Haschke, and
  Helge Ritter.
\newblock Using {{Tactile Sensing}} to {{Improve}} the {{Sample Efficiency}}
  and {{Performance}} of {{Deep Deterministic Policy Gradients}} for
  {{Simulated In-Hand Manipulation Tasks}}.
\newblock \emph{Frontiers in Robotics and AI}, 8:\penalty0 538773, 2021.
\newblock ISSN 2296-9144.

\bibitem[Romano et~al.(2011)Romano, Hsiao, Niemeyer, Chitta, and
  Kuchenbecker]{romano2011human}
Joseph~M Romano, Kaijen Hsiao, G{\"u}nter Niemeyer, Sachin Chitta, and
  Katherine~J Kuchenbecker.
\newblock Human-inspired robotic grasp control with tactile sensing.
\newblock \emph{IEEE Trans. on Robotics}, 27\penalty0 (6):\penalty0 1067--1079,
  2011.

\bibitem[Tahara et~al.(2010)Tahara, Arimoto, and Yoshida]{tahara2010dynamic}
Kenji Tahara, Suguru Arimoto, and Morio Yoshida.
\newblock Dynamic object manipulation using a virtual frame by a triple
  soft-fingered robotic hand.
\newblock In \emph{Proc. ICRA}, 2010.

\bibitem[Li et~al.(2012)Li, Haschke, Ritter, and Bolder]{li2012towards}
Qiang Li, Robert Haschke, Helge Ritter, and Bram Bolder.
\newblock Towards unknown objects manipulation.
\newblock \emph{IFAC Proceedings Volumes}, 45\penalty0 (22):\penalty0 289--294,
  2012.

\bibitem[Lach et~al.(2022)Lach, Lemaignan, Ferro, Ritter, and
  Haschke]{lachBioInspiredGraspingController2022}
Luca Lach, Severin Lemaignan, Francesco Ferro, Helge Ritter, and Robert
  Haschke.
\newblock Bio-{{Inspired Grasping Controller}} for {{Sensorized}} 2-{{DoF
  Grippers}}.
\newblock In \emph{2022 {{IEEE}}/{{RSJ International Conference}} on
  {{Intelligent Robots}} and {{Systems}} ({{IROS}})}, pages 11231--11237.
  {IEEE}, 2022.

\bibitem[Perrusquía et~al.(2019)Perrusquía, Yu, and
  Soria]{perrusquiaPositionForceControl2019}
Adolfo Perrusquía, Wen Yu, and Alberto Soria.
\newblock Position/force control of robot manipulators using reinforcement
  learning.
\newblock \emph{Industrial Robot: the international journal of robotics
  research and application}, 46\penalty0 (2):\penalty0 267--280, 2019.
\newblock ISSN 0143-991X, 0143-991X.

\bibitem[Luo et~al.(2019)Luo, Solowjow, Wen, Ojea, Agogino, Tamar, and
  Abbeel]{luoReinforcementLearningVariable2019}
Jianlan Luo, Eugen Solowjow, Chengtao Wen, Juan~Aparicio Ojea, Alice~M.
  Agogino, Aviv Tamar, and Pieter Abbeel.
\newblock Reinforcement {{Learning}} on {{Variable Impedance Controller}} for
  {{High-Precision Robotic Assembly}}.
\newblock In \emph{2019 {{International Conference}} on {{Robotics}} and
  {{Automation}} ({{ICRA}})}, pages 3080--3087. {IEEE}, 2019.

\bibitem[Todorov et~al.(2012)Todorov, Erez, and
  Tassa]{todorovMuJoCoPhysicsEngine2012a}
Emanuel Todorov, Tom Erez, and Yuval Tassa.
\newblock {{MuJoCo}}: {{A}} physics engine for model-based control.
\newblock In \emph{2012 {{IEEE}}/{{RSJ International Conference}} on
  {{Intelligent Robots}} and {{Systems}}}, pages 5026--5033. {IEEE}, 2012.

\bibitem[Johansson and Flanagan(2009)]{johansson2009coding}
Roland~S Johansson and J~Randall Flanagan.
\newblock Coding and use of tactile signals from the fingertips in object
  manipulation tasks.
\newblock \emph{Nature Reviews Neuroscience}, 10\penalty0 (5):\penalty0
  345--359, 2009.

\bibitem[Hsiao et~al.(2010)Hsiao, Chitta, Ciocarlie, and Jones]{Hsiao2010}
Kaijen Hsiao, Sachin Chitta, Matei Ciocarlie, and E.~Gil Jones.
\newblock Contact-reactive grasping of objects with partial shape information.
\newblock In \emph{IEEE/RSJ International Conference on Intelligent Robots and
  Systems}, 2010.

\bibitem[Patel et~al.(2018)Patel, Cox, and Correll]{Patel2018}
Radhen Patel, Rebecca Cox, and Nikolaus Correll.
\newblock Integrated proximity, contact and force sensing using
  elastomer-embedded commodity proximity sensors.
\newblock \emph{Autonomous Robots}, 42:\penalty0 1443--1458, 2018.

\bibitem[Merzic et~al.(2018)Merzic, Bogdanovic, Kappler, Righetti, and
  Bohg]{merzicLeveragingContactForces2018a}
Hamza Merzic, Miroslav Bogdanovic, Daniel Kappler, Ludovic Righetti, and
  Jeannette Bohg.
\newblock Leveraging {{Contact Forces}} for {{Learning}} to {{Grasp}}, 2018.

\bibitem[Wu et~al.(2019)Wu, Akinola, Varley, and Allen]{Wu2019}
Bohan Wu, Iretiayo Akinola, Jacob Varley, and Peter Allen.
\newblock Mat: Multi-fingered adaptive tactile grasping via deep reinforcement
  learning.
\newblock In \emph{3rd Conference on Robot Learning (CoRL 2019),}, 9 2019.

\bibitem[Kang et~al.(2023)Kang, Seong, Lee, and
  Shim]{kangVersatileDoorOpening2023}
Gyuree Kang, Hyunki Seong, Daegyu Lee, and D.~Hyunchul Shim.
\newblock A {{Versatile Door Opening System}} with {{Mobile Manipulator}}
  through {{Adaptive Position-Force Control}} and {{Reinforcement Learning}},
  2023.

\bibitem[Zhao et~al.(2020)Zhao, Queralta, and
  Westerlund]{zhaoSimtoRealTransferDeep2020}
Wenshuai Zhao, Jorge~Peña Queralta, and Tomi Westerlund.
\newblock Sim-to-{{Real Transfer}} in {{Deep Reinforcement Learning}} for
  {{Robotics}}: A {{Survey}}.
\newblock In \emph{2020 {{IEEE Symposium Series}} on {{Computational
  Intelligence}} ({{SSCI}})}, pages 737--744, 2020.

\bibitem[Ju et~al.(2022)Ju, Juan, Gomez, Nakamura, and
  Li]{juTransferringPolicyDeep2022}
Hao Ju, Rongshun Juan, Randy Gomez, Keisuke Nakamura, and Guangliang Li.
\newblock Transferring policy of deep reinforcement learning from simulation to
  reality for robotics.
\newblock \emph{Nature Machine Intelligence}, 4\penalty0 (12):\penalty0
  1077--1087, 2022.
\newblock ISSN 2522-5839.

\bibitem[Church and Lloyd(2021)]{churchTactileSimtoRealPolicy}
Alex Church and John Lloyd.
\newblock Tactile {{Sim-to-Real Policy Transfer}} via {{Real-to-Sim Image
  Translation}}.
\newblock \emph{5th Conference on Robot Learning (CoRL 2021)}, 2021.

\bibitem[Lin et~al.(2022)Lin, Lloyd, Church, and
  Lepora]{linTactileGymSimtoreal2022}
Yijiong Lin, John Lloyd, Alex Church, and Nathan~F. Lepora.
\newblock Tactile {{Gym}} 2.0: {{Sim-to-real Deep Reinforcement Learning}} for
  {{Comparing Low-cost High-Resolution Robot Touch}}, 2022.

\bibitem[Ward-Cherrier et~al.(2018)Ward-Cherrier, Pestell, Cramphorn, Winstone,
  Giannaccini, Rossiter, and Lepora]{ward-cherrierTacTipFamilySoft2018}
Benjamin Ward-Cherrier, Nicholas Pestell, Luke Cramphorn, Benjamin Winstone,
  Maria~Elena Giannaccini, Jonathan Rossiter, and Nathan~F. Lepora.
\newblock The {{TacTip Family}}: {{Soft Optical Tactile Sensors}} with
  {{3D-Printed Biomimetic Morphologies}}.
\newblock \emph{Soft Robotics}, 5\penalty0 (2):\penalty0 216--227, 2018.
\newblock ISSN 2169-5172, 2169-5180.

\bibitem[Lambeta et~al.(2020)Lambeta, Chou, Tian, Yang, Maloon, Most, Stroud,
  Santos, Byagowi, Kammerer, Jayaraman, and
  Calandra]{lambetaDIGITNovelDesign2020}
Mike Lambeta, Po-Wei Chou, Stephen Tian, Brian Yang, Benjamin Maloon,
  Victoria~Rose Most, Dave Stroud, Raymond Santos, Ahmad Byagowi, Gregg
  Kammerer, Dinesh Jayaraman, and Roberto Calandra.
\newblock {{DIGIT}}: {{A Novel Design}} for a {{Low-Cost Compact
  High-Resolution Tactile Sensor}} with {{Application}} to {{In-Hand
  Manipulation}}.
\newblock \emph{IEEE Robotics and Automation Letters}, 5\penalty0 (3):\penalty0
  3838--3845, 2020.
\newblock ISSN 2377-3766, 2377-3774.

\bibitem[Lepora et~al.(2022)Lepora, Lin, Money-Coomes, and
  Lloyd]{leporaDigiTacDIGITTacTipHybrid2022}
Nathan~F. Lepora, Yijiong Lin, Ben Money-Coomes, and John Lloyd.
\newblock {{DigiTac}}: {{A DIGIT-TacTip Hybrid Tactile Sensor}} for {{Comparing
  Low-Cost High-Resolution Robot Touch}}.
\newblock \emph{IEEE Robotics and Automation Letters}, 7\penalty0 (4):\penalty0
  9382--9388, 2022.
\newblock ISSN 2377-3766, 2377-3774.

\bibitem[Narang et~al.(2021{\natexlab{a}})Narang, Sundaralingam, Macklin,
  Mousavian, and Fox]{narangSimtoRealRoboticTactile2021}
Yashraj Narang, Balakumar Sundaralingam, Miles Macklin, Arsalan Mousavian, and
  Dieter Fox.
\newblock Sim-to-{{Real}} for {{Robotic Tactile Sensing}} via {{Physics-Based
  Simulation}} and {{Learned Latent Projections}}, 2021{\natexlab{a}}.

\bibitem[Narang et~al.(2021{\natexlab{b}})Narang, Sundaralingam, Van~Wyk,
  Mousavian, and Fox]{narangInterpretingPredictingTactile2021}
Yashraj~S. Narang, Balakumar Sundaralingam, Karl Van~Wyk, Arsalan Mousavian,
  and Dieter Fox.
\newblock Interpreting and {{Predicting Tactile Signals}} for the {{SynTouch
  BioTac}}, 2021{\natexlab{b}}.

\bibitem[Sferrazza and
  D’Andrea(2019)]{sferrazzaDesignMotivationEvaluation2019}
Carmelo Sferrazza and Raffaello D’Andrea.
\newblock Design, {{Motivation}} and {{Evaluation}} of a {{Full-Resolution
  Optical Tactile Sensor}}.
\newblock \emph{Sensors}, 19\penalty0 (4):\penalty0 928, 2019.
\newblock ISSN 1424-8220.

\bibitem[Sferrazza et~al.(2020)Sferrazza, Bi, and
  D'Andrea]{sferrazzaLearningSenseTouch2020}
Carmelo Sferrazza, Thomas Bi, and Raffaello D'Andrea.
\newblock Learning the sense of touch in simulation: A sim-to-real strategy for
  vision-based tactile sensing, 2020.

\bibitem[Sferrazza and
  D'Andrea(2021)]{sferrazzaSimtorealHighresolutionOptical2021}
Carmelo Sferrazza and Raffaello D'Andrea.
\newblock Sim-to-real for high-resolution optical tactile sensing: {{From}}
  images to {{3D}} contact force distributions, 2021.

\bibitem[Bi et~al.(2021)Bi, Sferrazza, and
  D’Andrea]{biZeroShotSimtoRealTransfer2021}
Thomas Bi, Carmelo Sferrazza, and Raffaello D’Andrea.
\newblock Zero-{{Shot Sim-to-Real Transfer}} of {{Tactile Control Policies}}
  for {{Aggressive Swing-Up Manipulation}}.
\newblock \emph{IEEE Robotics and Automation Letters}, 6\penalty0 (3):\penalty0
  5761--5768, 2021.
\newblock ISSN 2377-3766, 2377-3774.

\bibitem[Akkaya et~al.(2019)Akkaya, Andrychowicz, Chociej, Litwin, McGrew,
  Petron, Paino, Plappert, Powell, Ribas, et~al.]{akkaya2019solving}
Ilge Akkaya, Marcin Andrychowicz, Maciek Chociej, Mateusz Litwin, Bob McGrew,
  Arthur Petron, Alex Paino, Matthias Plappert, Glenn Powell, Raphael Ribas,
  et~al.
\newblock Solving rubik's cube with a robot hand.
\newblock \emph{arXiv preprint arXiv:1910.07113}, 2019.

\bibitem[Schulman et~al.(2017)Schulman, Wolski, Dhariwal, Radford, and
  Klimov]{schulmanProximalPolicyOptimization2017}
John Schulman, Filip Wolski, Prafulla Dhariwal, Alec Radford, and Oleg Klimov.
\newblock Proximal {{Policy Optimization Algorithms}}, 2017.

\bibitem[Mysore et~al.(2021)Mysore, Mabsout, Mancuso, and
  Saenko]{mysoreRegularizingActionPolicies2021}
Siddharth Mysore, Bassel Mabsout, Renato Mancuso, and Kate Saenko.
\newblock Regularizing {{Action Policies}} for {{Smooth Control}} with
  {{Reinforcement Learning}}, 2021.

\bibitem[Narvekar and Stone(2018)]{narvekarLearningCurriculumPolicies2018}
Sanmit Narvekar and Peter Stone.
\newblock Learning {{Curriculum Policies}} for {{Reinforcement Learning}},
  2018.

\bibitem[Narvekar(2020)]{narvekarCurriculumLearningReinforcement2020}
Sanmit Narvekar.
\newblock Curriculum learning for reinforcement learning domains.
\newblock \emph{Journal of Machine Learning Research}, 21, 2020.

\end{thebibliography}
